\documentclass[journal]{IEEEtran}
\usepackage{amsmath,amsfonts}
\usepackage{amsthm,amssymb}
\usepackage{mathrsfs}

\usepackage{algorithmic}
\usepackage{algorithm}
\usepackage{array}
\usepackage[caption=false,font=normalsize,labelfont=sf,textfont=sf]{subfig}
\usepackage{textcomp}
\usepackage{stfloats}
\usepackage{url}
\usepackage{verbatim}
\usepackage{graphicx}
\usepackage{cite}

\usepackage{soul}
\usepackage{bm}
\usepackage{booktabs}
\usepackage{txfonts}
\usepackage{booktabs}
\usepackage{multirow}
\usepackage{caption3}
\usepackage{xcolor}

\usepackage{listings}
\usepackage{pifont}
\usepackage{threeparttable}

\begin{document}
\title{RSCaMa: Remote Sensing Image Change Captioning with State Space Model
}

\author{Chenyang Liu, Keyan Chen, Bowen Chen, Haotian Zhang, Zhengxia Zou,~\IEEEmembership{Member,~IEEE}, \\and Zhenwei Shi$^*$,~\IEEEmembership{Senior Member,~IEEE}
\\
Beihang University


}



\maketitle

\begin{abstract}
Remote Sensing Image Change Captioning (RSICC) aims to describe surface changes between multi-temporal remote sensing images in language, including the changed object categories, locations, and dynamics of changing objects (e.g., added or disappeared). This poses challenges to spatial and temporal modeling of bi-temporal features. Despite previous methods progressing in the spatial change perception, there are still weaknesses in joint spatial-temporal modeling. To address this, in this paper, we propose a novel RSCaMa model, which achieves efficient joint spatial-temporal modeling through multiple CaMa layers, enabling iterative refinement of bi-temporal features. To achieve efficient spatial modeling, we introduce the recently popular Mamba (a state space model) with a global receptive field and linear complexity into the RSICC task and propose the Spatial Difference-aware SSM (SD-SSM), overcoming limitations of previous CNN- and Transformer-based methods in the receptive field and computational complexity. SD-SSM enhances the model's ability to capture spatial changes sharply. In terms of efficient temporal modeling, considering the potential correlation between the temporal scanning characteristics of Mamba and the temporality of the RSICC, we propose the Temporal-Traversing SSM (TT-SSM), which scans bi-temporal features in a temporal cross-wise manner, enhancing the model's temporal understanding and information interaction. Experiments validate the effectiveness of the efficient joint spatial-temporal modeling and demonstrate the outstanding performance of RSCaMa and the potential of the Mamba in the RSICC task. Additionally, we systematically compare three different language decoders, including Mamba, GPT-style decoder, and Transformer decoder, providing valuable insights for future RSICC research. The code will be available at \emph{\url{https://github.com/Chen-Yang-Liu/RSCaMa}}

\end{abstract}


\section{Introduction}
\IEEEPARstart{M}{ulti}-temporal remote sensing images captured by satellites provide rich data resources for surface dynamic monitoring \cite{BIFA,chen2023rsprompter}. Remote Sensing Image Change Captioning (RSICC) uses multi-temporal images of the same area to identify and describe surface changes in language, such as the changed object categories, the positional relationships between objects, and dynamics of changing objects (e.g., added or disappeared). It greatly enhances the readability of change interpretation information and has significant application in various fields, including land planning, disaster detection, and urban expansion research \cite{wang2024CD_review}.

RSICC involves remote sensing image processing and natural language generation. Current mainstream methods typically employ an encoder-decoder structure. In the encoder, the CNN or Vision Transformer (ViT) is used as the backbone to extract visual features from bi-temporal images, a well-designed ``neck" to enhance bi-temporal feature modeling and capture interesting changes features.
Finally, language models are used as the decoder to convert these visual features into captions. Most current research focuses on improving the neck, usually using CNN or Transformer as the basic component, and integrating attention mechanisms to facilitate the accurate perception of changes.

Hoxha \textit{et al.} \cite{RSICC_2} firstly explored the RSICC task and proposed two CNN-based bi-temporal fusion strategies. Liu \textit{et al.} \cite{RSICCformer} constructed a large dataset LEVIR-CC and benchmarked several methods such as DUDA \cite{robust_CC}, MCCFormer \cite{MCCformer}. Besides, they proposed a dual-branch Transformer to locate change areas and a Multistage Bitemporal Fusion (MBF) module to perform the fusion of multi-scale features. Chang \textit{et al.} \cite{RSICC_TIP2023} proposed a hierarchical self-attention module to locate change-related features. PSNet \cite{PSNet} leverages scale-aware reinforcement modules to enhance the perception of changed objects of varying sizes. PromptCC \cite{liu2023decoupling} uses an image-level classifier and a feature-level encoder to decouple the RSICC task. Besides, a multi-prompt learning strategy is proposed to exploit a large language model for captioning effectively. CT-Net \cite{RSICC_cai2023RS_interactive} use employs a Multi-Layer Adaptive Fusion (MAF) module to minimize the impact of irrelevant visual features and a Cross Gated-Attention (CGA) module to attend to multi-scale features during the caption-generation process. 

Although progress has been made, previous methods have primarily focused on enhancing spatial change perception capabilities, with weaknesses in joint spatial-temporal modeling. Besides, they are constrained by CNNs with limited receptive fields and Transformers with high complexity. Recently, State Space Models (SSMs) \cite{gu2021efficiently}, especially Mamba \cite{gu2023mamba}, as an emerging sequence model, have become increasingly popular due to their global receptive field and linear complexity. Mamba has demonstrated remarkable performance in many language and vision tasks \cite{Wang2024SSMSurvey, hao2024t_mamba} and is beginning to show promise in remote sensing field \cite{he2024pan_mamba,chen2024rsmamba,zhao2024rs_mamba,zhu2024samba}. 

Inspired by the above insight, we propose a novel RSCaMa model to achieve efficient joint spatial-temporal modeling through iterative bi-temporal feature refinement of multiple CaMa layers. Specifically, we introduce Mamba into the RSICC and propose Spatial Difference-aware SSM (SD-SSM) to enhance spatial change perception through differential features. Considering the potential correlation between the temporal scanning characteristics of Mamba and the temporality of the RSICC, to facilitate temporal modeling, we propose the Temporal-Traversing SSM (TT-SSM), which scans bi-temporal features in a temporal cross-wise manner, enhancing the model’s temporal understanding and information interaction. We experimentally confirm the effectiveness of RSCaMa's efficient joint spatial-temporal modeling and demonstrate the outstanding performance and potential of the Mamba in the RSICC. Additionally, we compare the effect of three language decoders: Mamba, GPT-style decoder with causal attention, and Transformer decoder with cross-attention, providing valuable insight for future RSICC research.

Our contributions can be summarized as follows:
\begin{itemize}
\item 
To facilitate efficient joint spatial-temporal modeling, we propose RCaMa, which introduces SSM into the RSICC and uses multiple CaMa layers to perform spatial change perception and temporal interaction iteratively.


\item 
We propose SD-SSM to enhance spatial change perception. To facilitate temporal modeling, we leverage the temporal scanning characteristics of Mamba and propose TT-SSM with a temporal cross-wise scanning manner.
\item 
The experiment demonstrates RSCaMa’s effective joint spatial-temporal modeling and Mamba’s potential in the RSICC. Besides, our evaluation of language decoding schemes provides valuable insights for future research.



\end{itemize}

\begin{figure*}
	\centering
	\includegraphics[width=1\linewidth]{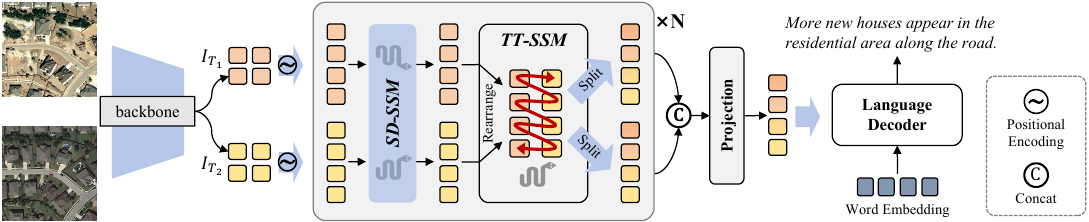}
	\caption{The illustration of the proposed RSCaMa, which consists of three main components: the backbone, multiple CaMa layers, and the language decoder. The CaMa layers play a pivotal role in facilitating efficient joint spatial-temporal modeling. Specifically, SD-SSM focuses on enhancing spatial change perception, while TT-SSM concentrates on temporal interaction.
 }
	\label{fig:RSCaMa}
\end{figure*}

\vspace{-10pt}
\section{Methodology}
In this section, we present an in-depth explanation of our proposed RSCaMa. We commence by introducing the foundation of the state space model. Following that, we offer a comprehensive overview of the RSCaMa model and delineate its principal constituents: SD-SSM and TT-SSM.

\vspace{-10pt}
\subsection{Preliminaries}
SSMs ~\cite{gu2021efficiently} represent a class of sequence models with linear complexity, inspired by linear time-invariant (LTI) continuous system mapping a one-dimensional function or sequence $x(t)\in \mathbb{R}$ to $y(t) \in \mathbb{R}$ through an implicit hidden state $h(t)\in \mathbb{R}^{N}$. The system can be mathematically expressed as follows:
\begin{align}
    &h'(t) = \mathbf{A}h(t)+\mathbf{B}x(t),&y(t) = \mathbf{C}h(t)
\end{align}
where $h'(t)$ refers to the time derivative of $h(t)$, ${A} \in \mathbb{R}^{N \times N}$ is the evolution parameter, $\mathbf{B} \in \mathbb{R}^{N \times 1}$ and $\mathbf{C} \in \mathbb{R}^{1 \times N}$ serve as the projection parameters.

To process discrete sequences and map the input sequence $\{x_1, x_2, ..., x_K\}$ to the output sequence $\{y_1, y_2, ..., y_K\}$, SSMs leverage zero-order hold (ZOH) discretization to convert continuous parameters $\mathbf{A}$ and $\mathbf{B}$ into their discrete counterparts $\Bar{\mathbf{A}}$ and $\Bar{\mathbf{B}}$. The transformation is achieved by incorporating a timescale parameter $\mathbf{\Delta} \in \mathbb{R}^{D}$ and can be formally defined as follows:
\begin{align}
    &\Bar{\mathbf{A}} = \mathbf{exp(\Delta A)},
    &\Bar{\mathbf{B}} =\mathbf{(\Delta A)}^{-1}\mathbf{(exp(\mathbf{\Delta} A)-I)\cdot \mathbf{\Delta} B} \approx \mathrm{\mathbf{\Delta}}B
\end{align}
Then, the discrete representation of the linear system can be formulated as follows:
\begin{align}
    &h_k = \Bar{\mathbf{A}}h_{k-1}+\Bar{\mathbf{B}}x_k,
    &y_k = \mathbf{C}h_k
\end{align}

Mamba \cite{gu2023mamba} breaks the LTI limitation of SSM by introducing a selective state space mechanism. It allows the matrices $\Bar{\mathbf{B}}$, ${\mathbf{C}}$, and $\mathbf{\Delta}$ to vary with the input sequences, enhancing selective information processing across sequences, enabling the model to be contextually aware of the input.

\vspace{-10pt}
\subsection{RSCaMa}
As depicted in Fig. \ref{fig:RSCaMa}, we present the overall architecture of the proposed RSCaMa. RSCaMa consists of three main components: the backbone, multiple CaMa layers, and the language decoder. Following prior research \cite{liu2023decoupling}, we employ the CLIP image encoder that has been image-text aligned as the backbone to extract bi-temporal features, which are then tokenized as $I_{T_i} \in \mathbb{R}^{L \times D} (i=1,2)$. $L$ and $D$ are the number and dimensions of tokens respectively. Subsequently, they are added with position embeddings and fed into multiple CaMa layers. The CaMa layers leverage the synergistic effect of SD-SSM and TT-SSM to facilitate efficient joint spatial-temporal modeling, with SD-SSM enhancing spatial change perception and TT-SSM concentrating on temporal interaction. The two token sequences undergo iterative spatial and temporal processing by multiple CaMa layers. Then they are input into a projection layer, bridging the gap between the visual and language domains. Finally, the language decoder utilizes the resulting embeddings to generate descriptive sentences. For the RSCaMa with $N$ CaMa layers, 
the flow of caption generation is expressed as follows:
\begin{align}
{I_{T_1}^{(0)}, I_{T_2}^{(0)}} &= \mathrm{\Phi_{pos}}(I_{T_1}, I_{T_2}) \\
{I_{T_1}^{(l)}, I_{T_2}^{(l)}} &= \mathrm{\Phi_{CaMa}^{(l)}}(I_{T_1}^{(l-1)}, I_{T_2}^{(l-1)}), l=(1,2,...N) \\
I_F &= \mathrm{\Phi_{Proj}}([I_{T_1}^{(N)};I_{T_2}^{(N)}]) \\
\mathrm{T_{cap}} &= \mathrm{\Phi_{Dec}}(I_F)
\end{align}
where [;] denotes feature concatenation operation on the channel dimension. $\mathrm{\Phi_{pos}}(\cdot)$ represents adding learnable positional embeddings. $\mathrm{\Phi_{Proj}}(\cdot)$ denotes a projection layer consisting of a linear layer and a residual convolution block.

\begin{figure}
	\centering
	\includegraphics[width=0.95\linewidth]{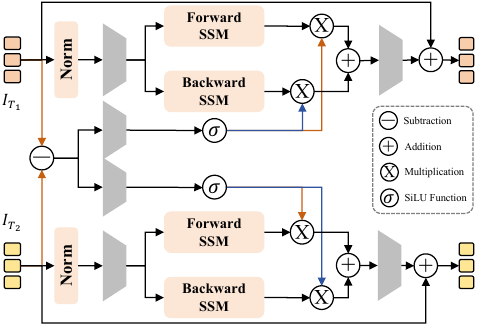}
	\caption{The structure of the SD-SSM. We multiply the differencing features and the output of bidirectional SSMs to improve the change-awareness.
 }
	\label{fig:SD_SSM}
 \vspace{-15pt}
\end{figure}

\subsection{SD-SSM}
Fig. \ref{fig:SD_SSM} presents the structure of our proposed SD-SSM. The standard Mamba is designed for one-dimensional sequences. Its causal scanning mechanism makes it unsuitable for spatial-aware understanding in visual tasks. Inspired by \cite{vim_mamba}, we perform Forward-SSM and Backward-SSM scans on flattened visual tokens to achieve bidirectional modeling, enhancing understanding of visual features. Furthermore, to improve the model's spatial perception of changes, we multiply the bi-temporal differencing features and the output of bidirectional SSMs to guide the model. Let's denote bi-temporal features input to SD-SSM as $I_{T_1}$ and $I_{T_2}$, the processing of SD-SSM can be represented as follows:
\begin{align}
I^o_{T_i} &= \mathrm{\Phi_{L}}(I^f_{T_i}*I_d + I^b_{T_i}*I_d)+I_{T_i}, i=1,2 \\
I_d &= \sigma(\mathrm{\Phi_{L}}(I_{T_2}-I_{T_1})) \\
I^f_{T_i} &= \mathrm{\Phi_{SSM}}(I^p_{T_i})  \\
I^b_{T_i} &= \mathrm{flip}(\mathrm{\Phi_{SSM}}(\mathrm{flip}(I^p_{T_i}))) \\
I^p_{T_i} &= \sigma(\mathrm{\Phi_{Dwc}(\Phi_{L}(\Phi_{Norm}}(I_{T_i}))))
\end{align}
where $I^o_{T_i}$ is the output, $\sigma$ is the SiLU activation function, $\mathrm{\Phi_{Dwc}}(\cdot),\mathrm{\Phi_{L}}(\cdot),\mathrm{\Phi_{Norm}}(\cdot)$ respectively represent depth-wise convolution, linear projection, and normalization. $\mathrm{flip} (\cdot)$ represents reversing the order of a sequence. 

\begin{figure}
	\centering
	\includegraphics[width=0.95\linewidth]{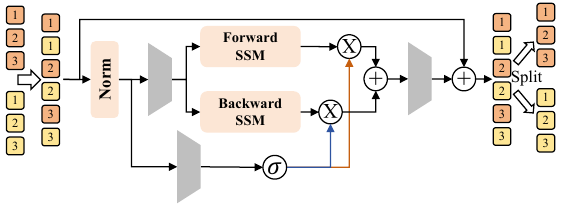}
	\caption{The structure of the TT-SSM, which rearranges two sequences in a bi-temporal token-wise interleaving manner to facilitate temporal modeling.
 }
	\label{fig:TT_SSM}
\vspace{-15pt}
\end{figure}

\subsection{TT-SSM}

SD-SSM enhances the model’s spatial change perception ability to capture changes sharply using differential features. Furthermore, we propose TT-SSM to facilitate temporal modeling. Considering the potential correlation between the temporal scanning characteristics of Mamba and the temporality of the RSICC, we propose a novel temporal cross-wise manner to scan bi-temporal token sequences. Fig. \ref{fig:TT_SSM} shows the structure of the TT-SSM. 
Specifically, for the two sequences $I_{T_1} \in \mathbb{R}^{L \times D} $ and $I_{T_2} \in \mathbb{R}^{L \times D}$ from SD-SSM, we rearrange them in a bi-temporal token-wise interleaving manner to construct sequences $I_{s}=[I_{T_1}^{(0)},I_{T_2}^{(0)},I_{T_1}^{(1)},I_{T_2}^{(1)}, I_{T_1}^{(2)},...] \in \mathbb{R}^{2L \times D}$ that effectively facilitate temporal information exchange. Finally, $I_{s}$ will be processed by bidirectional SSM and subsequently split into bi-temporal sequences in the original order. This elegant strategy facilitates bi-temporal information interaction. After that, the bi-temporal sequences will be sent to the next CaMa layer, which iteratively performs spatial and temporal processing.


\section{Experiments}
\subsection{Experimental Setting}
\subsubsection{Dataset}
LEVIR-CC \cite{RSICCformer} is a large-scale RSICC dataset comprising 10,077 pairs of images and 50,385 descriptive captions. These images capture changes over time in 20 distinct regions across Texas, USA, with a temporal range spanning from 5 to 15 years. 

\subsubsection{Evaluation Metrics}
We employ four quantitative metrics to measure the captioning accuracy \cite{Liu_2022}, including BLEU-N (N=1,2,3,4), ROUGE$_L$, METEOR, and CIDEr-D. Higher metric scores indicate better model performance. Besides, following \cite{liu2023decoupling}, we adopt an average metric $S^*_m$ as follows:
\begin{equation}
S^*_m = \frac{1}{4}\mathrm{(BLEU\text{-}4+ROUGE_L+METEOR+CIDEr\text{-}D)}
\end{equation}
\subsection{Training of RSCaMa}
\subsubsection{Objective Function}
We adopt a supervised learning way to train our model. The cross-entropy function is used to compute the loss between predicted captions and ground truth captions. The formulation of the captioning loss can be expressed as follows:
\begin{equation}
\mathcal{L}_{\text{cap}} = - \frac{1}{N} \sum_{n=1,v=1}^{N,V} y_{n}^{(v)} \log(p_{n}^{(v)})
\end{equation}
Here, $N$ represents the total number of word tokens, $V$ denotes the vocabulary size, $y_{n}$ denotes the one-hot vector representation of the $n$-th word in the reference sentence, and $p_{n}$ is probability vector of the predicted word. 

\subsubsection{Implementation Details}
We implemented all models using the PyTorch deep learning framework and trained them on the NVIDIA GTX 4090 GPU. During training, we use the Adam optimizer to optimize the model parameters. The dimension of the word vector is 768. The number of CaMa layers is set to 3. Our final RSCaMa uses Transformer decoder as the language decoder because of better performance.
\vspace{-15pt}
\subsection{Comparison to State-of-the-Art}
Currently, mainstream methods commonly utilize CNNs and Transformers as basic design components of the ``neck'' to process bi-temporal features extracted by the backbone. Methods like Capt-Rep-Diff \cite{robust_CC}, Capt-Att \cite{robust_CC}, Capt-Dual-Att \cite{robust_CC}, and DUDA \cite{robust_CC} combine CNNs and attention mechanisms to process bi-temporal features and employ LSTM for language decoding. MCCFormer-S \cite{MCCformer}, MCCFormer-D \cite{MCCformer}, RSICCFormer \cite{RSICCformer}, PSNet \cite{PSNet} and PromptCC \cite{liu2023decoupling} use Transformer as the core component to construct the neck and employ Transformer's decoder to generate captions. In this section, we compare RSCaMa with these state-of-the-art (SOTA) methods on model performance, number of parameters, and floating-point operations (FLOPs).

As shown in Table \ref{tab:Comparisons_other_methods}, RSCaMa demonstrates outstanding performance, particularly excelling in key metrics such as BLEU-4 and $S^*_m$. 
Specifically, compared to the latest PromptCC, our model shows +1.70\% on BLEU-4 and +1.11\% on $S^*_m$. This significant improvement is attributed to our innovative spatial-temporal modeling units: SD-SSM and TT-SSM. RSCaMa exploits their collaborative effect to promote efficient joint spatial-temporal modeling, which is different from previous methods that use CNN and Transformer to focus on spatial-aware modeling.
Besides, our RSCaMa maintains leading performance while having a low number of parameters and computational complexity. 
Experimental results also suggest that Mamba, as an emerging deep-learning model, has the potential to become a significant option in future model designs. Further exploration of combinations of Mamba with CNNs, Transformers, and Multilayer Perceptrons (MLPs) will be promising research to advance the RSICC technology.

\begin{table*}[ht] 
\vspace{-10pt}
\renewcommand{\arraystretch}{0.99}
\caption{Comparisons experiments on model performance, number of parameters, and floating-point operations (FLOPs).  Unlike previous methods relying on CNNs or Transformers, and focusing on spatial-aware modeling, our RSCaMa uses SD-SSM and TT-SSM to facilitate efficient joint spatial-temporal modeling.
}
\label{tab:Comparisons_other_methods}
\centering
\vspace{-8pt}
\begin{tabular}{cc|c c c c c c c| c | c c c}
	\toprule
	Type & Method & BLEU-1 & BLEU-2 & BLEU-3 & BLEU-4 & METEOR & ROUGE$_L$ & CIDEr-D & $S^*_m$ & Param. & FLOPs\\
	\midrule
	\multirow{4}{*}{\shortstack{CNN \\based}} & {Capt-Rep-Diff\cite{robust_CC}} & 72.90 & 61.98 & 53.62 & 47.41 & 34.47 & 65.64 & 110.57 & 64.52 & 73.21M & 19.82G\\
	 & {Capt-Att\cite{robust_CC}} & 77.64 & 67.40 & 59.24 & 53.15 & 36.58 & 69.73 & 121.22 & 70.17 & 73.60M & 19.90G\\
	 & {Capt-Dual-Att\cite{robust_CC}} & 79.51 & 70.57 & 63.23 & 57.46 & 36.56 & 70.69 & 124.42 & 72.28 & 75.58M & 20.03G\\
	 & {DUDA \cite{robust_CC}} & 81.44 & 72.22 & 64.24 & 57.79 & 37.15 & 71.04 & 124.32 & 72.58 & 80.31M & 20.28G\\
  \cline{1-2}
	 \multirow{5}{*}{\shortstack{Transformer\\based}} & {MCCFormer-S\cite{MCCformer}} & 79.90 & 70.26 & 62.68 & 56.68 & 36.17 & 69.46 & 120.39 & 70.68 & 162.55M & 25.09G\\
	 & {MCCFormer-D\cite{MCCformer}} & 80.42 & 70.87 & 62.86 & 56.38 & 37.29 & 70.32 & 124.44 & 72.11 & 162.55M & 25.09G\\
	 & {RSICCFormer\cite{RSICCformer}} & {84.72} & {76.27} & {68.87} & {62.77} & {39.61} & {74.12} & {134.12} & 77.65 & 172.80M & 27.10G\\
      & {PSNet\cite{PSNet}} & 83.86 & 75.13 & 67.89 & 62.11 & 38.80 & 73.60 & 132.62 & 76.78 & 319.76M & 13.78G\\
      & {PromptCC\cite{liu2023decoupling}} & 83.66 & 75.73 & 69.10 & 63.54 & 38.82 & 73.72 & 136.44 & 78.13 & 408.58M & 19.88G\\
	\midrule
 SSM based & {RSCaMa} & \textbf{85.79} & \textbf{77.99} & \textbf{71.04} & \textbf{65.24} & \textbf{39.91} & \textbf{75.24} & \textbf{136.56} & \textbf{79.24} & {176.90M} & {13.03G}\\
 
	\bottomrule
\end{tabular}
\vspace{-8pt}
\end{table*}

\begin{table*}[ht] 
\renewcommand{\arraystretch}{0.99}
\vspace{-6pt}
\caption{Ablation studies on the SD-SSM and TT-SSM. TT-SSM* means that two sequences are concatenated on length instead of the proposed arrangement.
}
\label{tab:Ablation_CaMa}
\vspace{-8pt}
\centering
\begin{tabular}{c c c c | c c c c c c c | c}
	\toprule
	Method & SD-SSM & TT-SSM & TT-SSM* & BLEU-1 & BLEU-2 & BLEU-3 & BLEU-4 & METEOR & ROUGE$_L$ & CIDEr-D & $S^*_m$\\
	\midrule
	{Baseline} &\ding{55} &\ding{55} &\ding{55} & 85.14 & 77.23 & 70.37 & 64.75 & 39.40 & 74.38 & 135.28 & 78.45 \\
	{--} &\ding{52} &\ding{55} &\ding{55} & 85.75 & 77.85 & 70.85 & 65.01 & 39.67 & 74.90 & 135.91 & 78.87 \\
	{RSCaMa} &\ding{52} &\ding{52} &\ding{55} & \textbf{85.79} & \textbf{77.99} & \textbf{71.04} & \textbf{65.24} & \textbf{39.91} & \textbf{75.24} & \textbf{136.56} & \textbf{79.24} \\
 {--} &\ding{52} &\ding{55} &\ding{52}  
 & 85.64 & 77.77 & 70.72 & 64.86 & 39.44 & 74.51 & 134.97 & 78.45 \\
	\bottomrule
\end{tabular}
\vspace{-12pt}
\end{table*}

\vspace{-16pt}
\subsection{Ablation Studies}
\subsubsection{CaMa Layers} 
The CaMa layers consist of SD-SSM and TT-SSM and play a pivotal role in facilitating efficient joint spatial-temporal modeling. 
In Table \ref{tab:Ablation_CaMa}, the structure of the baseline is similar to SD-SSM, but the difference lies in using self-features rather than differential features to multiply with the output of bidirectional SSM. TT-SSM* means we concatenate two sequences on length instead of adopting the proposed token-wise cross-scanning manner.
Experimental results validate that SD-SSM and TT-SSM significantly contribute to improving change captioning performance. This is because SD-SSM effectively enhances the model's understanding of spatial changes by multiplying differential features with the output of bidirectional SSM, while TT-SSM promotes effective temporal interaction of bi-temporal features through a token-wise cross-scanning manner. 
The comparison between TT-SSM and TT-SSM* demonstrates that the effectiveness of TT-SSM stems from the proposed cross-scanning manner rather than from the addition of extra bidirectional SSM layers.

We also conducted experiments on the number of CaMa layers. As shown in Table \ref{tab:Ablation_camalayer_number}, our method achieves the best performance when we set the number of CaMa layers to 3. 

\begin{table} 
\vspace{-10pt}
\renewcommand{\arraystretch}{0.99}
\caption{Parametric experiments on the number of CaMa layers.}
\vspace{-8pt}
\label{tab:Ablation_camalayer_number}
\centering
\begin{tabular}{c |c c c c |c}
	\toprule
	Num. & BLEU-4 & METEOR & ROUGE$_L$ & CIDEr-D & $S^*_m$\\
	\midrule
	{2} & 64.78 & 39.77 & 74.75 & 136.16 & 78.87 \\
	{3} & \textbf{65.24} & \textbf{39.91} & \textbf{75.24} & {136.56} & \textbf{79.24} \\
 	{4} & 64.77 & 39.79 & 74.86 & \textbf{136.72} & 79.04 \\
	\bottomrule
\end{tabular}
\vspace{-10pt}
\end{table}


\begin{table} 
\renewcommand{\arraystretch}{0.99}
\vspace{-5pt}
\caption{Ablation studies on three language decoders. 
}
\vspace{-8pt}
\label{tab:Ablation_decoder}
\centering
\begin{tabular}{c |c c c c |c}
	\toprule
	Method & BLEU-4 & METEOR & ROUGE$_L$ & CIDEr-D & $S^*_m$\\
	\midrule
	{Mamba} & 63.13 & 37.83 & 72.61 & 127.26 & 75.21 \\
	{GPT-style} & 64.04 & 39.36 & 74.10 & 134.64 & 78.04 \\
	{Trans-Dec} & \textbf{65.24} & \textbf{39.91} & \textbf{75.24} & \textbf{136.56} & \textbf{79.24} \\
	\bottomrule
\end{tabular}
\vspace{-20pt}
\end{table}

\subsubsection{Language Decoder}
RSICC require transforming visual change into language. The language decoders adopted by previous methods have evolved with the development of language models, from recurrent neural networks to Transformers. As a novel language model, the performance of Mamba in cross-modal understanding is yet to be explored. In Table \ref{tab:Ablation_decoder}, we compare three language decoders: the Mamba language model, the Transformer decoder with cross-attention, and the GPT-style decoder with causal attention only. In our experiments, for the first two methods, we prefix the visual embeddings to the language decoders to generate words one by one. For the Transformer decoder with cross-attention, visual embeddings are integrated into the decoding process as keys (K) and values (V) of the cross-attention mechanism. The experimental results in Table \ref{tab:Ablation_decoder} demonstrate that the Transformer decoder with cross-attention mechanism performs best in transforming visual to textual information, surpassing the simple visual embedding prefix strategy (i.e., Mamba and GPT-style decoder), thus proving the superiority of cross-attention in cross-modal understanding. In the future, it is promising to explore combining Mamba with novel attention mechanisms rather than the simple visual prefix strategy to enhance its cross-modal understanding ability when generating each word.

\begin{figure*}
	\centering
        \vspace{-10pt}
	\includegraphics[width=1\linewidth]{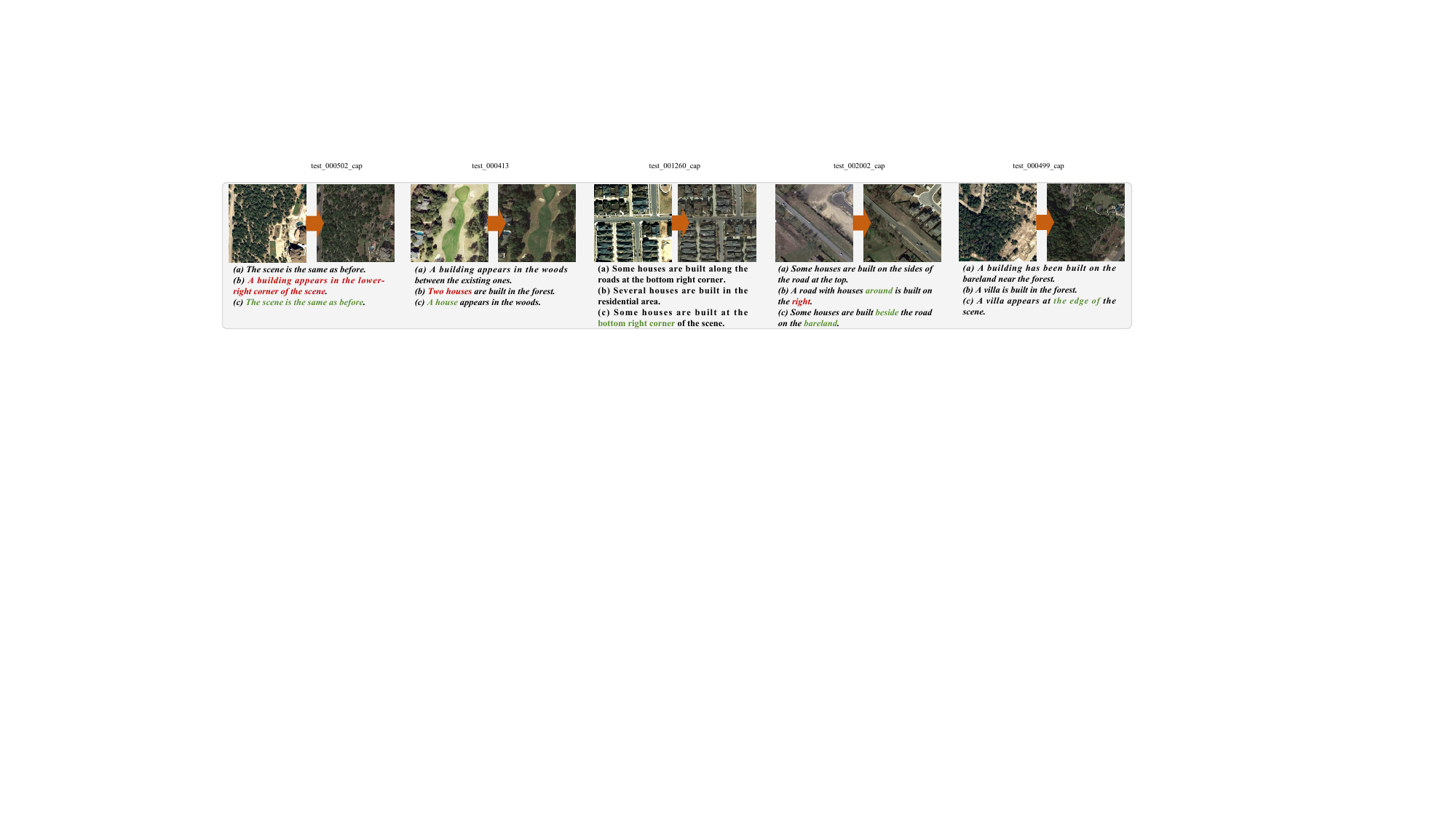}
         \vspace{-20pt}
	\caption{Captioning results on the LEVIR-CC dataset. Sentence (a) is one of the five ground-truth sentences. Sentence (b) is from the baseline, while (c) is from our RSCaMa. More accurate and detailed words are marked in green. Red words are not accurate. 
 }
\label{fig:cap_result}
\vspace{-18pt}
\end{figure*}

\begin{figure}
	\centering
	\includegraphics[width=1\linewidth]{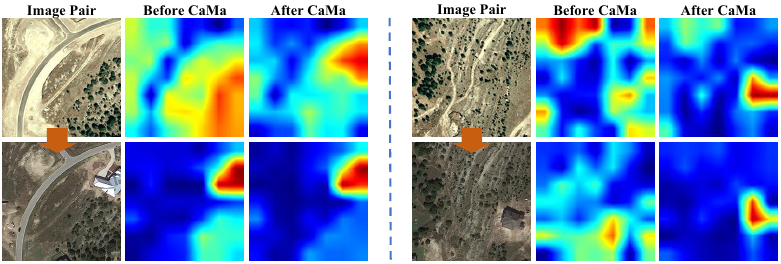}
         \vspace{-20pt}
	\caption{Visualization of features before and after CaMa layer processing. 
 }
\label{fig:feat_vis}
\vspace{-18pt}
\end{figure}

\vspace{-10pt}
\subsection{Qualitative Results}
Fig. \ref{fig:cap_result} showcases some captioning results. 
We can observe that our RSCaMa demonstrates a remarkable ability to accurately distinguish changes in the scenes, as illustrated by the first image pair. Furthermore, our RSCaMa excels in identifying the categories of changing objects, their locations, and the relationships between objects. For instance, in the second image pair, our RSCaMa accurately identifies the number of the changed houses, while in the third image pair, it correctly recognizes that the location of changed buildings is in the lower right corner. 

In Fig. \ref{fig:feat_vis}, we visualise bi-temporal features before and after CaMa layer processing. Experimental results show that our proposed CaMa layer helps extract discriminative features that reveal changing regions. 


\vspace{-10pt}
\section{Conclusion}
Different from previous methods that use CNN and Transformer and mainly focus on spatial change perception modeling, we propose RCaMa to achieve efficient joint spatial-temporal modeling
through iterative bi-temporal feature refinement of multiple
CaMa layers. Specifically, we introduce Mamba (a state space model) into the RSICC task and propose SD-SSM and TT-SSM. SD-SSM uses differential features to enhance spatial change perception, while TT-SSM promotes bitemporal interactions in a token-wise cross-scanning manner. Experiments demonstrate the superiority of RSCaMa over existing methods, highlighting Mamba's potential in RSICC. Besides, our systematic evaluation of three language decoding schemes provides valuable insights for future research.

\ifCLASSOPTIONcaptionsoff
\newpage
\fi

\vspace{-5pt}
\bibliographystyle{IEEEtran}
\vspace{-5pt}
\bibliography{papers.bib}

\begin{thebibliography}{10}
\providecommand{\url}[1]{#1}
\csname url@samestyle\endcsname
\providecommand{\newblock}{\relax}
\providecommand{\bibinfo}[2]{#2}
\providecommand{\BIBentrySTDinterwordspacing}{\spaceskip=0pt\relax}
\providecommand{\BIBentryALTinterwordstretchfactor}{4}
\providecommand{\BIBentryALTinterwordspacing}{\spaceskip=\fontdimen2\font plus
\BIBentryALTinterwordstretchfactor\fontdimen3\font minus \fontdimen4\font\relax}
\providecommand{\BIBforeignlanguage}[2]{{%
\expandafter\ifx\csname l@#1\endcsname\relax
\typeout{** WARNING: IEEEtran.bst: No hyphenation pattern has been}%
\typeout{** loaded for the language `#1'. Using the pattern for}%
\typeout{** the default language instead.}%
\else
\language=\csname l@#1\endcsname
\fi
#2}}
\providecommand{\BIBdecl}{\relax}
\BIBdecl

\bibitem{BIFA}
H.~Zhang, H.~Chen, C.~Zhou, K.~Chen, C.~Liu, Z.~Zou, and Z.~Shi, ``Bifa: Remote sensing image change detection with bitemporal feature alignment,'' \emph{IEEE Transactions on Geoscience and Remote Sensing}, vol.~62, pp. 1--17, 2024.

\bibitem{chen2023rsprompter}
K.~Chen, C.~Liu, H.~Chen, H.~Zhang, W.~Li, Z.~Zou, and Z.~Shi, ``Rsprompter: Learning to prompt for remote sensing instance segmentation based on visual foundation model,'' \emph{arXiv preprint arXiv:2306.16269}, 2023.

\bibitem{wang2024CD_review}
L.~Wang, M.~Zhang, X.~Gao, and W.~Shi, ``Advances and challenges in deep learning-based change detection for remote sensing images: A review through various learning paradigms,'' \emph{Remote Sensing}, vol.~16, no.~5, p. 804, 2024.

\bibitem{RSICC_2}
G.~Hoxha, S.~Chouaf, F.~Melgani, and Y.~Smara, ``Change captioning: A new paradigm for multitemporal remote sensing image analysis,'' \emph{IEEE Transactions on Geoscience and Remote Sensing}, pp. 1--1, 2022.

\bibitem{RSICCformer}
C.~Liu, R.~Zhao, H.~Chen, Z.~Zou, and Z.~Shi, ``Remote sensing image change captioning with dual-branch transformers: A new method and a large scale dataset,'' \emph{IEEE Transactions on Geoscience and Remote Sensing}, vol.~60, pp. 1--20, 2022.

\bibitem{robust_CC}
D.~H. Park, T.~Darrell, and A.~Rohrbach, ``Robust change captioning,'' in \emph{2019 IEEE/CVF International Conference on Computer Vision (ICCV)}, 2019, pp. 4623--4632.

\bibitem{MCCformer}
Y.~Qiu, S.~Yamamoto, K.~Nakashima, R.~Suzuki, K.~Iwata, H.~Kataoka, and Y.~Satoh, ``Describing and localizing multiple changes with transformers,'' in \emph{2021 IEEE/CVF International Conference on Computer Vision (ICCV)}, 2021, pp. 1951--1960.

\bibitem{RSICC_TIP2023}
S.~Chang and P.~Ghamisi, ``Changes to captions: An attentive network for remote sensing change captioning,'' \emph{IEEE Transactions on Image Processing}, 2023.

\bibitem{PSNet}
C.~Liu, J.~Yang, Z.~Qi, Z.~Zou, and Z.~Shi, ``Progressive scale-aware network for remote sensing image change captioning,'' in \emph{IGARSS 2023-2023 IEEE International Geoscience and Remote Sensing Symposium}.\hskip 1em plus 0.5em minus 0.4em\relax IEEE, 2023, pp. 6668--6671.

\bibitem{liu2023decoupling}
C.~Liu, R.~Zhao, J.~Chen, Z.~Qi, Z.~Zou, and Z.~Shi, ``A decoupling paradigm with prompt learning for remote sensing image change captioning,'' \emph{IEEE Transactions on Geoscience and Remote Sensing}, 2023.

\bibitem{RSICC_cai2023RS_interactive}
C.~Cai, Y.~Wang, and K.-H. Yap, ``Interactive change-aware transformer network for remote sensing image change captioning,'' \emph{Remote Sensing}, vol.~15, no.~23, p. 5611, 2023.

\bibitem{gu2021efficiently}
A.~Gu, K.~Goel, and C.~R{\'e}, ``Efficiently modeling long sequences with structured state spaces,'' \emph{arXiv preprint arXiv:2111.00396}, 2021.

\bibitem{gu2023mamba}
A.~Gu and T.~Dao, ``Mamba: Linear-time sequence modeling with selective state spaces,'' \emph{arXiv preprint arXiv:2312.00752}, 2023.

\bibitem{Wang2024SSMSurvey}
X.~Wang, S.~Wang, Y.~Ding, Y.~Li, W.~Wu, Y.~Rong, W.~Kong, J.~Huang, S.~Li, H.~Yang, Z.~Wang, B.~Jiang, C.~Li, Y.~Wang, Y.~Tian, and J.~Tang, ``State space model for new-generation network alternative to transformers: A survey,'' 2024.

\bibitem{hao2024t_mamba}
J.~Hao, L.~He, and K.~F. Hung, ``T-mamba: Frequency-enhanced gated long-range dependency for tooth 3d cbct segmentation,'' \emph{arXiv preprint arXiv:2404.01065}, 2024.

\bibitem{he2024pan_mamba}
X.~He, K.~Cao, K.~Yan, R.~Li, C.~Xie, J.~Zhang, and M.~Zhou, ``Pan-mamba: Effective pan-sharpening with state space model,'' \emph{arXiv preprint arXiv:2402.12192}, 2024.

\bibitem{chen2024rsmamba}
K.~Chen, B.~Chen, C.~Liu, W.~Li, Z.~Zou, and Z.~Shi, ``Rsmamba: Remote sensing image classification with state space model,'' \emph{arXiv preprint arXiv:2403.19654}, 2024.

\bibitem{zhao2024rs_mamba}
S.~Zhao, H.~Chen, X.~Zhang, P.~Xiao, L.~Bai, and W.~Ouyang, ``Rs-mamba for large remote sensing image dense prediction,'' \emph{arXiv preprint arXiv:2404.02668}, 2024.

\bibitem{zhu2024samba}
Q.~Zhu, Y.~Cai, Y.~Fang, Y.~Yang, C.~Chen, L.~Fan, and A.~Nguyen, ``Samba: Semantic segmentation of remotely sensed images with state space model,'' \emph{arXiv preprint arXiv:2404.01705}, 2024.

\bibitem{vim_mamba}
L.~Zhu, B.~Liao, Q.~Zhang, X.~Wang, W.~Liu, and X.~Wang, ``Vision mamba: Efficient visual representation learning with bidirectional state space model,'' \emph{arXiv preprint arXiv:2401.09417}, 2024.

\bibitem{Liu_2022}
C.~Liu, R.~Zhao, and Z.~Shi, ``Remote sensing image captioning based on multi-layer aggregated transformer,'' \emph{IEEE Geoscience and Remote Sensing Letters}, pp. 1--1, 2022.

\end{thebibliography}

\end{document}